%% file: paper_v2.tex
\documentclass[10pt,twocolumn,letterpaper]{article}

\usepackage[algorithms]{wacv}

\usepackage[T1]{fontenc}
\usepackage{enumitem}
\usepackage{booktabs}
\usepackage{graphicx}
\usepackage{amsmath,amssymb}
\usepackage[expansion=false]{microtype}

\newtheorem{definition}{Definition}

\newcommand{\rlvr}{RLVR}
\newcommand{\rans}{r_{\text{ans}}}
\newcommand{\rord}{r_{\text{ord}}}
\newcommand{\rfmt}{r_{\text{fmt}}}

\definecolor{wacvblue}{rgb}{0.21,0.49,0.74}
\usepackage[pagebackref,breaklinks,colorlinks,allcolors=wacvblue]{hyperref}

\def\wacvPaperID{2942}
\def\confName{WACV}
\def\confYear{2027}

\title{What Do Verifiable Rewards Teach Video--Language Models About Time?
A Controlled Multi-Model Study}

\author{Avyay Sadhu\thanks{Equal contribution.}\\
University of Colorado Boulder, United States
\and
Patrick Cooper\footnotemark[1]\\
University of Colorado Boulder, United States
}

\begin{document}
\maketitle

\begin{abstract}
\noindent
Reinforcement learning from verifiable rewards (\rlvr{}) has produced large
reasoning gains in language models, and verifiable video benchmarks make
the recipe directly applicable to causal--temporal video question answering.
We present a controlled study of what \rlvr{} actually teaches
video--language models about temporal structure. We fine-tune four open
models with group relative policy optimization (GRPO) under three data
recipes using annotated synthetic questions, self-supervised real-video pretext
tasks, and a one-to-one mixture, plus a targeted real-video augmentation arm.
We evaluate the model--recipe grid on 105 CLEVRER questions with one correct
option and on two real-video transfer sets. The 8B ordered gain is 4.8 points
(95\% video-bootstrap CI $[-5.6,15.1]$), while blind accuracy increases
10.5 points ($[0.9,20.6]$). The 4B model gains an accuracy advantage for ordered over shuffled
frames. On the real-video sets, the 8B synthetic-only checkpoint loses 26.7 and 25.2
points, with smaller regressions under mixed and augmented recipes.
In a separate scene-disjoint, single-gold Qwen3-VL-8B experiment,
supervised fine-tuning (SFT) and GRPO each complete 1,200 updates at three
fixed seeds. GRPO minus SFT is $-3.89$ points in ordered accuracy
(nominal 97.5\% video-bootstrap CI $[-7.36,-0.45]$) and $-2.78$ points
in ordered-minus-shuffled accuracy ($[-6.60,0.87]$).
Transfer differences are sensitive to parseability. Intervals quantify
test-video uncertainty conditional on the fitted runs. The results show
why benchmark accuracy, transfer, and presented-order sensitivity need
to be evaluated separately.
\end{abstract}

\begin{figure*}[t]
\centering
\includegraphics[width=\textwidth]{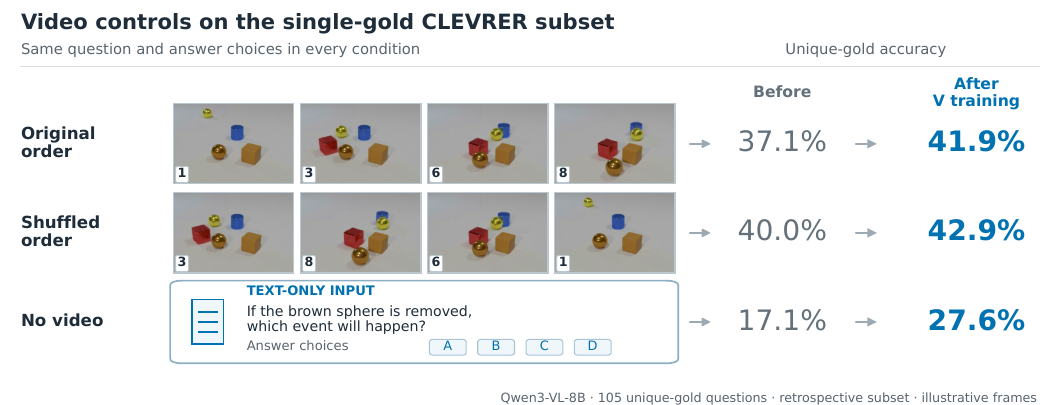}
\caption{\textbf{Controls on the valid unique-gold subset.}
An actual scene illustrates matched inputs. Four of eight reconstructed,
cropped frames receive the same display-only tone and color adjustment.
Source frames, question and choices are unchanged. Scores
aggregate 105 questions from 86 videos, not the pictured item. For the
8B model, ordered accuracy is 37.1\%$\to$41.9\%, shuffled accuracy
40.0\%$\to$42.9\%, and blind accuracy 17.1\%$\to$27.6\%.
Table~\ref{tab:rlvf-main} reports the corresponding intervals.
Shuffling retains content and state cues.}
\label{fig:overview}
\end{figure*}

\section{Introduction}
\label{sec:intro}

Film editors have long exploited the fact that the meaning of a sequence
of images depends on their order. The Kuleshov effect and the montage
tradition demonstrate that viewers construct causal and narrative
interpretations from juxtaposition rather than from shots in isolation
\cite{kuleshov1974,eisenstein1949,bordwell2017}.
For video--language models, comparing ordered and shuffled versions of
the same frames tests whether presented frame order improves accuracy.
It does not measure all temporal understanding. Physical state
cues can reveal chronology after shuffling, and some questions remain
answerable from the unordered frame set. Prior work has documented weak
dependence on video sequence information and single-frame shortcuts
\cite{buch2022revisiting,lei2023single}.

Reinforcement learning from verifiable rewards has recently produced
significant reasoning improvements in language models
\cite{shao2024deepseekmath,deepseekr1,lambert2024tulu}. When the
correctness of an output can be checked programmatically, policy
optimization can climb the reward without human preference labels.
Causal--temporal video QA offers programmatically checkable answers, and
synthetic benchmarks such as CLEVRER \cite{yi2020clevrer} also provide
ground-truth event orders. This motivates comparing answer-reward training
with training on self-supervised temporal tasks, measuring benchmark
accuracy, transfer, and the advantage of receiving frames in their
original order.

This paper asks, with a controlled experimental design, what \rlvr{}
training actually changes in a video--language model. Three properties are
distinguishable, namely (i) in-domain answer accuracy, (ii) transfer of that
accuracy to real-video questions, and (iii) the paired
ordered-versus-shuffled accuracy difference on held-out items. Our
historical design crosses four open models with three data recipes using
verified synthetic data, unverified real video with self-supervised
pretext rewards, and a one-to-one mixture, plus a targeted fourth arm that
augments the verified data with real-video verifiable questions.
Completed cells are evaluated under a common ordered/shuffled/blind
protocol with video-level bootstrap confidence intervals and test-set
fingerprints.

On the 105 unique-gold CLEVRER questions, in-domain gains vary across
models. The 8B model improves with no visual input, while the 4B verified
checkpoint gains an advantage from ordered frames. The 8B synthetic-only
checkpoint also shows substantial regressions on both real-video transfer
sets. Mixed and augmented recipes have smaller observed regressions.

We present a multi-model, multi-recipe empirical study of verifiable-reward
reinforcement learning for video reasoning. Accuracy gains alone
should not be interpreted as temporal understanding. No-frame and
shuffled controls help audit what changed. The ordered/shuffled/blind
evaluation protocol, the reward suite with its anti-gaming guards, and
the full pipeline are included in the accompanying reproducibility
package. Figure~\ref{fig:overview} summarizes the control audit.
We also report a matched SFT--GRPO comparison on a separate single-gold
CLEVRER population.

\section{Related Work}
\label{sec:related}

\paragraph{Temporal order and montage.}
That ordered juxtaposition creates meaning absent from individual images
is foundational in film theory. Kuleshov's editing experiments
\cite{kuleshov1974} and Eisenstein's montage writings \cite{eisenstein1949}
document that interpretation is constructed \emph{between} shots, a view
elaborated by modern film scholarship \cite{bordwell2017}. We adopt this
observation as motivation for a diagnostic based on comparing ordered and
shuffled frames, rather than as a novel claim. Interpreting this
comparison in QA requires checking whether shuffling removes information
needed to answer the question. In machine learning, shuffling and single-frame controls have
shown that video--language benchmarks are often solvable without temporal
information \cite{buch2022revisiting,lei2023single}. Those studies
characterize \emph{benchmarks and pretrained models}, whereas we ask
whether targeted reinforcement learning can \emph{change} the property,
extending the control to paired per-item statistics with clustered
uncertainty estimates measured before and after training.

\paragraph{Causal--temporal video benchmarks.}
CLEVRER \cite{yi2020clevrer} isolates collision causality with descriptive,
explanatory, predictive, and counterfactual questions over synthetic
physics scenes and provides exact ground truth, making it a natural source
of verifiable rewards. NExT-QA \cite{xiao2021nextqa} targets temporal and
causal questions on real web video. MVBench \cite{li2024mvbench}
assembles twenty temporal subtasks from eleven
video datasets, VCRBench \cite{sarkar2025vcrbench}
evaluates long-form causal reasoning via clip reordering, and ChainReaction
\cite{parmar2025chainreaction} structures causal-why QA around explicit
causal chains. We train on CLEVRER-derived items (verified arm) and
evaluate out-of-domain transfer on NExT-QA-derived temporal items and a
broad MVBench subset, using a split construction intended to keep test videos
disjoint from training.

\paragraph{Reinforcement learning from verifiable rewards.}
Group relative policy optimization (GRPO)
\cite{shao2024deepseekmath} optimizes a policy against
group-normalized rewards without a value network and underlies recent
reasoning-focused post-training \cite{deepseekr1}. RLVR
\cite{lambert2024tulu} formalizes training against programmatically
checkable answers. Reward source is distinct from optimizer choice.
PPO \cite{schulman2017ppo} and REINFORCE-style leave-one-out baselines
\cite{ahmadian2024reinforce} provide related policy-gradient alternatives.
Our RL experiments use GRPO.
Applications to multimodal models have focused on
answer-level rewards for image reasoning. Systematic evidence about what
such training changes in \emph{video} models beyond benchmark scores
is scarce. Our study fills this gap with controls (shuffled and blind
inputs, out-of-domain sets) designed to measure performance changes.

\paragraph{Self-supervised video pretext tasks.}
Temporal pretext tasks such as ordering shuffled clips, verifying arrow of
time, and anticipating future frames have a long history in video
representation learning. Our unverified arm ports these ideas to policy
optimization. Rewards are computed from the video itself (e.g.,
reconstructing the true position of frames), so unlabeled web videos
\cite{farre2024finevideo} become training signal without external answer
annotations.

\section{Method}
\label{sec:method}

\subsection{Policy optimization}
\label{sec:grpo}

Let $\pi_\theta$ be a vision--language model that maps a prompt $x$
(interleaved instruction text and $K$ video frames) to a text completion
$y$. For each training prompt we sample a group of $G$ completions
$y_1,\dots,y_G \sim \pi_\theta(\cdot \mid x)$, score each with a scalar
reward $R(y_g, x)$ (Sec.~\ref{sec:rewards}), and apply GRPO
\cite{shao2024deepseekmath}. The advantage of $y_g$ is its reward
standardized within the group,
\begin{equation}
A_g \;=\; \frac{R(y_g,x) - \operatorname{mean}_{g'} R(y_{g'},x)}
                {\operatorname{std}_{g'} R(y_{g'},x) + \epsilon},
\label{eq:grpo}
\end{equation}
with numerical stabilizer $\epsilon>0$. Constant-reward groups have zero
advantage. The policy gradient is taken on token log-probabilities
weighted by $A_g$. We use $G{=}8$ completions per prompt, LoRA adapters
\cite{hu2022lora} (rank 32, $\alpha{=}64$, dropout 0.05, attention
projections) on a frozen bf16 base model, learning rate $10^{-5}$,
per-device batch size~1 and eight gradient-accumulation steps. Recovered trainer states confirm
1{,}200 optimizer steps for all 11 reported runs. Sampling uses a colocated
vLLM engine \cite{kwon2023vllm} with per-token truncated importance-sampling
correction between the sampler and trainer log-probabilities. Each QA prompt carries up to $K{=}8$ frames sampled uniformly from the
clip, capped at $401{,}408$ pixels per frame. Pretext tasks use task-arranged
frames, prefixes, or candidates. Under the intended common protocol, after training, adapters are merged
into the base weights and evaluation is performed on the merged model by
the same serving stack used for base models.

\subsection{Verifiable rewards}
\label{sec:rewards}

The reward is a weighted sum $R = \rans + \rord + 0.2\,\rfmt$ of three
programmatically checkable terms, routed by task. Multiple-choice, direction and anticipation use
answer reward. Ordering uses order reward. Both use format reward.
Inactive terms are zero. V's multiple-choice records contain no order
labels, so an order-weight setting does not provide order supervision.

\begin{definition}[Answer reward]
For multiple-choice items with target letter $a^\star$ and choice set $C$,
$\rans(y)=\mathbf{1}[\operatorname{parse}(y)=a^\star]$.
The parser prefers a delimited answer block, with fallback to explicit
answer/option markers or a single-letter response, and accepts only
members of $C$. Unparseable answers score zero.
\end{definition}

\begin{definition}[Order reward]
For items with a ground-truth event or clip order $\sigma^\star$,
$\rord(y)=(\tau(\hat\sigma,\sigma^\star)+1)/2$, where the stated order
$\hat\sigma$ is parsed from an answer block or explicit order markers.
The modified Kendall score counts pairs involving omitted elements as
discordant. Unparseable orders score zero.
\end{definition}

Instruction-echo and explicit-order parsing guards limit reward gaming.
Only terms varying within a completion group affect its normalized
advantage. Supplementary Appendix~I preserves the complete equations,
parser rules, modified Kendall score and guards. These historical
vLLM/format-plus-answer/order recipes are distinct from the fresh native,
answer-only SFT--GRPO comparison in Supplementary Appendix~H.

\subsection{Three data recipes}
\label{sec:recipes}

\paragraph{Verified (V).} The historical converter maps
CLEVRER questions \cite{yi2020clevrer} to one answer letter, retaining
the last correct choice when several are true. The matched SFT--GRPO
experiment instead uses questions with exactly one correct choice.
The converter's multiple-choice records carry no event-order labels, so their active reward
is $\rans + 0.2\,\rfmt$. Optional VCRBench ordering augmentation is
not enabled by the provided Slurm workflow. The builder enforces
scene-disjoint splits.

\paragraph{Unverified (U).} Self-supervised pretext items over
FineVideo \cite{farre2024finevideo}. The recovered training manifest has
4{,}860 tasks from 1{,}215 videos.
Tasks include temporal jigsaw (recover the true position of shuffled
frame groups), arrow-of-time direction, and anticipation (select
which candidate frame follows a prefix). Jigsaw uses $\rord$. Direction and anticipation use
$\rans$, with ground truth computed from the video itself. Thus active
reward terms differ with task mix. ``Unverified'' denotes the absence of
external answer annotations, not programmatically checkable rewards.
Recovered train and holdout manifests have no item or video overlap.
Counts and source-metadata limits appear in the supplement.

\paragraph{Mixed (V+U).} A one-to-one interleaving of the two sources at the
prompt level, holding total optimization steps fixed. Relative to V,
this changes both the source/task distribution and exposure to synthetic
verified prompts.

\paragraph{Verified+Real (V+R).} A targeted arm, run on Qwen3-VL-8B,
configured with V's reward terms and hyperparameters but augmenting the
CLEVRER items with a pool of up to $4{,}000$ multiple-choice questions from
the NExT-QA train split (causal and temporal types), whose labels make
them verifiable. This tests transfer under real-video augmentation while
retaining verifiable rewards. As this arm is configured to train on NExT-QA, we treat the temporal
stress set (built from the validation split, Sec.~\ref{sec:protocol}) as
near-domain rather than out-of-domain for it. MVBench remains a transfer
test for every arm.

\subsection{Models}
\label{sec:models}

We train Qwen3-VL-8B, Qwen3-VL-4B \cite{bai2025qwen3vl}, Qwen2.5-VL-7B \cite{bai2025qwen25vl},
and Gemma-3-12B \cite{gemma3_2025}, four variants spanning three model
series, two vendors, and a 3$\times$ parameter range. All four receive the V recipe.
The three Qwen models receive all three recipes, and Qwen3-VL-8B
additionally receives the V+R arm (the Gemma U/V+U cells
did not complete within our compute budget and are omitted rather than
partially reported). Three further candidates were excluded for cause,
documented in the supplementary material. Gemma-3-4B scores 0.3\% on the archived selected-option proxy, below
its uniform-choice reference and has only partial recovered training records.
InternVL3.5-14B / VideoLLaMA3-7B could not be served faithfully by the
sampling stack (dynamic-tiling and unsupported-architecture
incompatibilities, respectively). The available evaluation archive
contains 15 CLEVRER, 11 stress, and 15 MVBench cells. Missing cells are
not imputed.

\input{rlvf_e3_main_methods.tex}

\section{Evaluation protocol}
\label{sec:protocol}

\paragraph{Test sets.} The historical in-domain manifest contains 300
CLEVRER counterfactual questions, with 105 having one correct choice,
99 having two, and 96 having three. QA accuracy is evaluated on the 105
unique-gold questions from 86 videos. The supplement also reports
agreement with the converter's selected target letter over all 300 items.
This selected-option metric uses only the last correct letter for each
question. Two out-of-domain sets probe transfer to real video.
The \emph{temporal stress} set contains $n{=}300$ multiple-choice items
built from NExT-QA \cite{xiao2021nextqa} temporal questions. The
\emph{broad} set contains $n{=}306$ items sampled evenly from nine
MVBench \cite{li2024mvbench} subtasks spanning six source datasets
(SSv2, Moments-in-Time, FunQA, Perception Test, MoVQA scenes, and VLN).
We exclude MVBench's five CLEVRER-derived subtasks, whose videos share
our verified training domain, along with license-gated,
frame-folder-format, and segment-bounded subtasks
(scoping rationale in the supplementary material). Every evaluation
embeds a SHA-256 fingerprint of
its item identifiers in its output. All evaluations of a given test set
share one fingerprint. This checks identifier alignment, not the identity
of question text, labels, or frame contents.

\paragraph{Input conditions.} Each item is evaluated under three
conditions, namely \textbf{ordered} (frames in true temporal order),
\textbf{shuffled} (the same frames, deterministically permuted per item),
and \textbf{blind} (no frames). Ordered accuracy measures task
performance. The paired ordered$-$shuffled difference measures the net
accuracy advantage of presented frame order. Blind accuracy measures
performance without visual input, including text-only inference and
response-format effects.

\begin{definition}[Order sensitivity]
For items $i = 1,\dots,n$ with per-item correctness $c^{\mathrm{ord}}_i,
c^{\mathrm{shuf}}_i \in \{0,1\}$, order sensitivity is
$\Delta_{\mathrm{ord}} = \tfrac1n \sum_i (c^{\mathrm{ord}}_i -
c^{\mathrm{shuf}}_i)$, with uncertainty from a bootstrap that resamples
\emph{videos} (clusters), not items.
\end{definition}

\paragraph{Target semantics and labels.} The unique-gold subset has
19 A, 25 B, 33 C, and 28 D answers (25\% uniform choice, 31.4\%
test-majority reference). The 300-item selected-option metric has a 52\%
D-label reference. Its label-balanced diagnostics average over those selected
targets. The unique-gold subset was selected retrospectively.

\paragraph{Statistics.} For the historical model--recipe grid, intervals
are 95\% percentile bootstrap CIs with source-video resampling on
CLEVRER and stress. Archived MVBench
intervals use task--video groups. Clustering addresses within-group
dependence. Greedy decoding was requested
(temperature $0$) for the reported primary evaluations. Effective backend settings were not retained. The harness refuses items without
a well-formed multiple-choice prompt and reports error and empty-response
counts and structured-format (scaffold) scores alongside accuracy. No
reported evaluation recorded a generation error or empty response.
Scaffold scores measure response structure. Answer-parse rates are
unavailable for the historical evaluations. Intervals for the 41
evaluation cells are exploratory and pointwise, without adjustment for
multiple comparisons.

\paragraph{Reproducibility.} Tables and figures are regenerated from
archived evaluation artifacts. Data splits and frame shuffling use fixed
seeds. A run-signature guard checks configuration compatibility on resume.
It does not fingerprint dataset contents. The accompanying package also
contains a CPU-only audit that verifies item hashes and summary means,
reports label distributions, and computes paired checkpoint changes where
video-group metadata is recoverable.\footnote{During camera-ready
reconstruction, OpenAI Codex was used for repository and data reconstruction,
experimental planning, software, orchestration and execution, audits and
analysis code, figure preparation, and manuscript drafting and revision.
Gold labels derive from source annotations. Reported metrics are computed
by scripts from retained model outputs.} Configurations are fixed per arm.
Software dependencies are not fully version-pinned. Recovered MVBench
metadata reproduces its archived intervals and supplies video clusters.
Training records confirm completed runs and observed source counts.

\section{Results}
\label{sec:results}

\input{rlvf_historical_primary_table.tex}

Table~\ref{tab:rlvf-main} reports the unique-gold CLEVRER subset.
Transfer tables appear in Supplementary Appendices~C--D, and transfer
and order-sensitivity plots appear in Appendices~C and G.

\subsection{In-domain gains vary across models}
\label{sec:f1}

On the 105 valid unique-gold counterfactual questions, 8B ordered accuracy changes
37.1\%$\to$41.9\% ($+4.8$ points, paired 95\% CI $[-5.6,15.1]$).
The analogous V-minus-Base
changes are $+19.0$ points for 4B ($[7.8,30.4]$), $-2.9$ for
Qwen2.5-VL-7B ($[-10.0,4.7]$), and $+22.9$ for Gemma
($[12.6,33.0]$). These exploratory intervals concern test-video
sampling conditional on the fitted checkpoints.

The complete paired-change table appears in Supplementary Appendix~G.

\subsection{A blind control reveals non-visual gains}
\label{sec:f2}

On the valid subset, 8B blind accuracy changes 17.1\%$\to$27.6\%
($+10.5$ points, CI $[0.9,20.6]$). Blind changes for 4B,
Qwen2.5-VL-7B, and Gemma are $+20.0$, $+15.2$, and $+5.7$ points,
respectively. These gains occur without visual input. Formatting,
answer priors, and text-only inference remain possible contributors.
On MVBench, Qwen2.5-VL-7B blind accuracy still rises from
$0.150$ to $0.395$--$0.431$ under the three recipes.

\subsection{Real-video transfer depends on model and training recipe}
\label{sec:f3}

On the real-video stress set (Supplementary Appendix~C, NExT-QA panel),
verified-only
training severely degraded Qwen3-VL-8B. The model falls from
$0.707$ to
$0.440$ ordered accuracy, a $26.7$-point regression (paired 95\% CI $[-32.9,-20.3]$), with losses distributed broadly across items (95
items
lost against 15 gained) and across all three temporal question categories
(per-category breakdown in the supplementary material). The degradation
also appears on the MVBench
broad subset (Supplementary Appendix~D and Appendix~C, MVBench panel). The same
checkpoint falls from $0.621$ to $0.369$ ($-25.2$ points, paired CI
$[-31.6,-18.9]$), declining on every
one of the nine subtasks, most severely on character ordering ($0.853
\to 0.118$), following verified multiple-choice training. The
transfer changes differ across models. Qwen3-VL-4B
gains out of domain on both sets ($+2.7$ stress, $+8.2$ broad) and
Gemma-3-12B
gains $+22.2$ on the broad set. Qwen2.5-VL-7B also loses 4.3 points
on the stress set (paired CI $[-8.6,-0.3]$). Thus the severe failure is
model-dependent, and in-domain improvement does not guarantee transfer.
For Qwen3-VL-8B, mixing reduces the observed stress regression to 2.7
points (paired CI $[-7.1,+1.7]$). Across the completed mixed cells,
the largest observed decline is 3.9 points on MVBench.

\paragraph{Real-video augmentation.} On Qwen3-VL-8B, V+R reaches
$0.598$ on MVBench versus $0.621$ at base and $0.369$ under V (paired
base-to-V+R change $-2.3$ points, CI $[-6.5,+2.0]$). It reaches $0.697$ on
the near-domain NExT-QA stress set versus $0.707$ at base (paired change
$-1.0$ points, CI $[-4.9,+2.7]$). In this run, V+R has a much smaller regression while retaining verifiable
rewards.
Adding real questions or mixing pretext items changes source diversity,
task composition, and synthetic exposure together at a fixed optimizer-step
budget. Paired checkpoint intervals for all three test sets are provided in
the supplementary material, using the recovered MVBench video mapping.

The targeted paired comparison directly favors V+R over V by
25.7 points on the stress set (95\% CI $[19.9,31.4]$) and 22.9 on
MVBench ($[16.9,29.1]$). Mixed also exceeds V on both sets. These retrospective video-bootstrap intervals apply to the fitted
checkpoints, without adjustment for multiple comparisons.

\subsection{Little consistent advantage from presented frame order}
\label{sec:f4}

Across the unique-gold CLEVRER and real-video evaluations, 38 of 41 presented-order-advantage
intervals include zero. The point estimates include 22 positive, 17 negative,
and two zero. The 4B V checkpoint has a positive 12.4-point advantage
on CLEVRER (CI $[2.7,22.0]$), with a Base-to-V increase of 16.2 points
($[2.9,29.2]$). The two other positive intervals are 4B Base and U on
MVBench. These comparisons concern individual fitted checkpoints and
are not adjusted for multiple testing.

The median interval width is 10.2 points (range 4.7--19.3). An interval
containing zero is not an equivalence result. A frame permutation retains
objects, states, and physical cues, while net accuracy can conceal
opposing item-level changes. The contrast measures the advantage of
\emph{presented} order.

\subsection{Fresh controlled SFT--GRPO comparison}
\label{sec:fresh-results}

Table~\ref{tab:fresh-e3-primary} reports the two CLEVRER co-primary
contrasts, averaged over three fixed seeds. Base is evaluated once per
population. Supplementary Appendix~H gives all arm accuracies and
parse/format diagnostics.

\begingroup
\setlength{\tabcolsep}{3pt}
\input{tables/e3_primary_contrasts.tex}
\endgroup

\begin{figure*}[t]
\centering
\includegraphics[width=\textwidth]{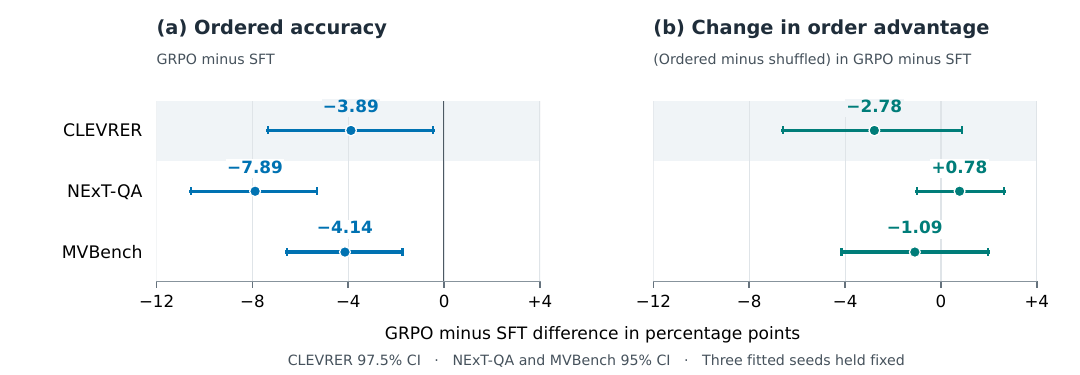}
\caption{\textbf{Fresh fixed-run SFT--GRPO contrasts.} Points and whole-video
intervals summarize the matched experiment.
CLEVRER uses nominal 97.5\% co-primary intervals. NExT-QA and MVBench
use exploratory pointwise 95\% intervals. Intervals condition on the
three fitted training seeds. Transfer accuracy includes parse failures.}
\label{fig:fresh-e3-contrasts}
\end{figure*}

\paragraph{In-domain comparison.}
On the new CLEVRER population, ordered accuracy is 60.67\% for Base,
71.33\% for SFT and 67.44\% for GRPO. The GRPO-minus-SFT ordered
contrast is $-3.89$ points (97.5\% CI $[-7.36,-0.45]$), with negative
differences at each fixed seed. The change in ordered-minus-shuffled
accuracy is $-2.78$ points ($[-6.60,0.87]$). Its interval spans zero.
All CLEVRER responses are parseable,
so unparseable responses do not contribute to this accuracy gap.
Strict answer-format compliance nevertheless differs, with 0.44\% for
GRPO's ordered inputs versus 100\% for SFT.

\paragraph{Transfer and output validity.}
On the retained NExT-QA and MVBench subsets, GRPO-minus-SFT ordered
accuracy is $-7.89$ points (pointwise 95\% CI $[-10.57,-5.32]$) and
$-4.14$ ($[-6.56,-1.74]$), respectively.
GRPO's ordered parse rates are 89.67\% and 94.77\%, versus 99.89\%
and 100\% for SFT. On the outcome-selected common-valid sets, the
corresponding contrasts are $+0.75$ points ($[-0.87,2.48]$) and
$-0.58$ ($[-2.71,1.52]$). These descriptive comparisons use
outcome-selected denominators. Figure~\ref{fig:fresh-e3-contrasts}
shows the unconditional contrasts, which include parse failures.

\section{Discussion}
\label{sec:discussion}

\paragraph{What the controls identify.} Blind improvements show that
post-training changes behavior even without visual input. Answer priors,
formatting, and text-only inference are possible contributors. Order
sensitivity measures the accuracy difference between two input conditions.
A reward that assigns credit \emph{only} when order is decision-relevant
is a natural next step. Testing it would require examples whose correct
response changes under a valid reversal or reordering of events, while
controlling frame content and ensuring that both sequences depict
plausible events. Merely shuffling frames does not establish that the
original answer becomes unrecoverable. Such a purpose-built test is
outside the present study.

\paragraph{Practical guidance.} For practitioners applying \rlvr{} to
video models, our results support reporting no-frame and shuffled controls
alongside headline gains and inspecting answer-label balance. Report
answer-parse rates separately from scaffold scores, together with accuracy
conditional on a valid parse. Monitor a
video-disjoint real-video validation set during post-training rather than
relying on in-domain gains. The historical 8B checkpoint has substantial
transfer regressions and an uncertain gain on the valid in-domain subset. Real-video augmentation
and mixing are promising ways to reduce these regressions, but require
validation for each model and target domain. For applications that require
event order,
include audited examples where content alone cannot recover the relevant
sequence, and report uncertainty on paired changes.

\paragraph{Historical limitations.} Our largest historical model is 12B
parameters and our optimizer budget is 1{,}200 steps per run, with GRPO,
LoRA, and one training run per completed model--recipe cell. Video-bootstrap intervals quantify uncertainty from the sampled test
videos, not variation across training seeds. That grid lacks matched
supervised fine-tuning on the same data. The matched SFT--GRPO comparison
covers one architecture and its declared subsets. The incomplete grid,
small test sets (105 valid in-domain and 300--306 transfer items), one augmentation source and one model for V+R,
and confounded data recipes restrict generalization. Eight sampled
frames can miss relevant motion. The shuffle test is not an item-level
order-dependence audit. Procedural clip-order and causal-chain evaluations
\cite{sarkar2025vcrbench,parmar2025chainreaction} are complementary.
Historical primary evaluation records contain correctness flags but lack
predicted letters, full generations, and answer-parse flags. The annotated
training contents and complete evaluation-to-weight-digest bindings are
unavailable, limiting verification of task mix, training labels, and
train/test separation.

\input{rlvf_e3_main_limitations.tex}

\section{Conclusion}
\label{sec:conclusion}

We conducted a controlled, multi-model study of reinforcement learning
from verifiable rewards for causal--temporal video reasoning. On the
unique-gold CLEVRER subset, gains vary across models, blind
accuracy increases, and the 4B verified checkpoint gains a positive
presented-order advantage. Real-video transfer degrades severely for the
8B synthetic-only checkpoint, with smaller regressions under augmented
and mixed recipes. The matched SFT--GRPO experiment favors SFT in ordered
CLEVRER accuracy for the fixed runs, while the order-advantage contrast
spans zero and transfer gaps depend on parseability. Together, these
results support evaluating benchmark accuracy, transfer, and presented-order
sensitivity separately when assessing verifiable-reward training.

\section*{Acknowledgments}
We thank Alvaro Velasquez for his guidance and support throughout this work.
We acknowledge computing support from CU Boulder's Alpine cluster, funded
by CU Boulder, CU Anschutz, Colorado State University, and NSF awards
OAC-2201538 and OAC-2322260.

\end{document}

%% file: rlvf_e3_main_methods.tex
\paragraph{Matched SFT--GRPO experiment.}
We compare a fixed
Qwen3-VL-8B-Instruct Base checkpoint with SFT and GRPO at seeds $0,1,2$.
Base is evaluated once per population.
Each of the six training chains completed $1{,}200$ updates on a frozen
common source pool. Evaluation uses a new single-gold,
scene-disjoint CLEVRER test population and retained NExT-QA/MVBench
subsets. Ordered and deterministically
shuffled conditions share eight frames bounded to $131{,}072$ pixels
each. Blind omits frames. All conditions request answer-only greedy
decoding capped at 64 new tokens. The two CLEVRER co-primary endpoints
are GRPO minus SFT in ordered accuracy and in ordered-minus-shuffled
accuracy, averaging contrasts over the three fixed seeds. These endpoints
use paired whole-video resampling with 10{,}000 PCG64 draws and nominal 97.5\%
intervals conditional on these runs.
Supplementary Appendix~H provides methods, exposure/resource
accounting and training-audit details.

%% file: rlvf_historical_primary_table.tex
\begin{table*}[t]\centering
\caption{\textbf{In-domain results} (CLEVRER unique-gold subset, $n{=}105$ from 86 videos, pointwise 95\% video-bootstrap CIs). Each question has one correct option. Accuracy is reported for frames in true order, shuffled frames and no frames (blind). $\Delta$ is the ordered-accuracy change from that model's base.}
\label{tab:rlvf-main}
\small\begin{tabular}{llcccc}\toprule
Model & Recipe & Ordered & $\Delta$ & Shuffled & Blind \\ \midrule
Qwen3-VL-8B & Base & 0.371\,[0.274,0.472] & -- & 0.400\,[0.311,0.491] & 0.171\,[0.105,0.242] \\
 & Verified & 0.419\,[0.327,0.515] & +0.048 & 0.429\,[0.336,0.525] & 0.276\,[0.200,0.355] \\
 & Verified+Real & 0.467\,[0.369,0.564] & +0.095 & 0.486\,[0.388,0.580] & 0.238\,[0.160,0.320] \\
 & Unverified & 0.429\,[0.333,0.524] & +0.057 & 0.400\,[0.311,0.495] & 0.200\,[0.128,0.276] \\
 & Mixed & 0.429\,[0.333,0.524] & +0.057 & 0.448\,[0.349,0.543] & 0.171\,[0.106,0.240] \\
\midrule
Qwen3-VL-4B & Base & 0.162\,[0.093,0.235] & -- & 0.200\,[0.126,0.275] & 0.048\,[0.010,0.092] \\
 & Verified & 0.352\,[0.262,0.442] & +0.190 & 0.229\,[0.149,0.314] & 0.248\,[0.165,0.333] \\
 & Unverified & 0.505\,[0.404,0.602] & +0.343 & 0.495\,[0.402,0.588] & 0.000\,[0.000,0.000] \\
 & Mixed & 0.448\,[0.356,0.535] & +0.286 & 0.429\,[0.333,0.523] & 0.229\,[0.154,0.309] \\
\midrule
Qwen2.5-VL-7B & Base & 0.667\,[0.570,0.759] & -- & 0.676\,[0.585,0.764] & 0.162\,[0.096,0.231] \\
 & Verified & 0.638\,[0.552,0.721] & -0.029 & 0.667\,[0.579,0.752] & 0.314\,[0.224,0.408] \\
 & Unverified & 0.676\,[0.582,0.766] & +0.010 & 0.667\,[0.579,0.755] & 0.257\,[0.181,0.337] \\
 & Mixed & 0.648\,[0.561,0.730] & -0.019 & 0.695\,[0.615,0.775] & 0.276\,[0.198,0.358] \\
\midrule
Gemma-3-12B & Base & 0.076\,[0.028,0.132] & -- & 0.105\,[0.049,0.168] & 0.181\,[0.112,0.255] \\
 & Verified & 0.305\,[0.221,0.394] & +0.229 & 0.305\,[0.216,0.400] & 0.238\,[0.157,0.324] \\
\bottomrule
\end{tabular}\end{table*}

%% file: tables/e3_primary_contrasts.tex
\begin{table*}[t]\centering
\caption{Matched Qwen3-VL-8B study. SFT/GRPO are arithmetic means of three fixed training seeds. CLEVRER GRPO-minus-SFT co-primary endpoints use nominal 97.5\% whole-test-video intervals for the family of two, conditional on the fitted seeds. Seed 0/1/2 and sample SD are descriptive percentage-point spreads.}
\label{tab:fresh-e3-primary}
\small\begin{tabular}{llllll}\toprule
Endpoint & Estimate [interval] & Seed 0 & Seed 1 & Seed 2 & Seed SD \\ \midrule
Ordered accuracy & -3.89\,[-7.36,-0.45] & -4.33 & -4.33 & -3.00 & 0.77 \\
Ordered-minus-shuffled & -2.78\,[-6.60,0.87] & -1.67 & -4.67 & -2.00 & 1.64 \\
\bottomrule\end{tabular}\end{table*}

%% file: rlvf_e3_main_limitations.tex
\paragraph{Matched-study limitations.}
The fresh study covers one 8B architecture and three fixed training seeds per
method. Video-bootstrap intervals condition on these runs and quantify
test-video uncertainty. Matched updates control optimizer-step counts.
Available token counters
have different scopes, while FLOPs and parameter deltas remain unmeasured.
Transfer tests reuse previously inspected populations, restricting conclusions
to these subsets. One fixed permutation per item probes presented-order
sensitivity. The study does not include an item-level audit of whether
shuffling removes information needed to answer each question.